\documentclass[review]{article}

\usepackage{lineno,hyperref}
\modulolinenumbers[5]

\usepackage{epsfig,array,amsmath,amssymb,float}
\usepackage{xspace,calc,multirow,theorem}
\usepackage[applemac]{inputenc}
\usepackage[]{graphicx}

\DeclareMathAlphabet{\mathbfeul}{U}{eur}{b}{n}
\DeclareMathAlphabet{\matheul}{U}{eur}{m}{n}
\DeclareMathAlphabet{\mathbfeus}{U}{eus}{b}{n}
\DeclareMathAlphabet{\matheus}{U}{eur}{m}{n}
\DeclareMathAlphabet{\mathbfzc}{OML}{pzc}{b}{n}
\DeclareMathAlphabet{\mathzc}{OT1}{pzc}{m}{it}
\DeclareMathAlphabet{\mathag}{OT1}{pag}{m}{n}
\DeclareMathAlphabet{\mathbfeuf}{U}{euf}{b}{n}
\DeclareMathAlphabet{\matheuf}{U}{euf}{m}{n}
\newcommand {\tr} {\rm\scriptscriptstyle T}

\newcommand{\be}[1]{\begin{equation} #1 \end{equation}}
\newcommand{\bea}[1]{\begin{eqnarray} #1 \end{eqnarray}}

\newcommand{\bMat}[1]{\begin{bmatrix}#1\end{bmatrix}}

\newcommand {\eqrefn}{Eq.~\eqref}

\newcommand{\Exp}[1]{\operatorname{E}\left[#1\right]}  

  \newcommand{\rbf}{\mathbf{r}}
  
  \newcommand{\xbf}{\mathbf{x}}
\newcommand{\ybf}{\mathbf{y}}

\newcommand{\epsilonbf}{\boldsymbol{\epsilon}}

\newcommand{\Sigmabf}{\boldsymbol{\Sigma}}

\newcommand{\Abf}{\mathbf{A}}  
 \newcommand{\Ebf}{\mathbf{E}} \newcommand{\Fbf}{\mathbf{F}}
  \newcommand{\Ibf}{\mathbf{I}}
  \newcommand{\Lbf}{\mathbf{L}}
\newcommand{\Mbf}{\mathbf{M}}  
\newcommand{\Pbf}{\mathbf{P}} \newcommand{\Qbf}{\mathbf{Q}} \newcommand{\Rbf}{\mathbf{R}}
\newcommand{\Sbf}{\mathbf{S}}  \newcommand{\Ubf}{\mathbf{U}}
\newcommand{\Vbf}{\mathbf{V}} \newcommand{\Wbf}{\mathbf{W}} \newcommand{\Xbf}{\mathbf{X}}
\newcommand{\Ybf}{\mathbf{Y}}  

\newcommand{\zeros}{\boldsymbol{0}}
\newcommand{\Rbb}{\mathbb{R}}
{\theoremstyle{break}\newtheorem{property}{Property}\theoremheaderfont{\normalfont\bfseries}
\newcommand {\vlightrule}{\kern1ex\vrule width0.2pt\kern1ex}

\def\endkeywords{\vspace{0.6em}\par\if@twocolumn\else\endquotation\fi
    \normalsize\rm}

\usepackage[]{setspace}
\doublespacing

\bibliographystyle{plain}

\begin{document}

\title{Deconstructing  Principal Component Analysis Using a Data Reconciliation Perspective}
\author{
        \begin{tabular}[t]{cccc}
     Shankar Narasimhan\footnote{Corresponding authors. E-mail: {\tt naras@iitm.ac.in\, niravbhatt@iitm.ac.in,}},  & Nirav Bhatt$^{\star}$
        \end{tabular}
	\\
	Systems $\&$ Control Group, Department of Chemical Engineering, \\ Indian Institute of Technology Madras, Chennai - 600036, India
}
\maketitle

\begin{abstract}
Data reconciliation (DR) and Principal Component Analysis (PCA) are two popular data analysis techniques in process industries. Data reconciliation is used to obtain accurate and consistent estimates of variables and parameters from erroneous measurements. PCA is primarily used as a method for reducing the dimensionality of high dimensional data and as a preprocessing technique for denoising measurements.  These techniques have been developed and deployed independently of each other.  The primary purpose of this article is to elucidate the close relationship between these two seemingly disparate techniques.  This leads to a unified framework for applying PCA and DR. Further, we show  how the two techniques can be deployed together in a collaborative and consistent manner to process data. The framework has been extended to deal with  partially measured systems and to incorporate partial knowledge available about the process model. 
\end{abstract}



\section{Introduction }
Data Reconciliation (DR) is a technique that was proposed in the early 1950s to derive accurate and consistent estimates of process variables and parameters from noisy measurements. This technique has been refined and developed over the past fifty years. Several books and book chapters have been written on this and related techniques \cite{romagnolibook99,Madron97,NarasimhanJ00,Hodouin10,Bagajewicz01}.  The technique is now an integral part of simulation software packages such as ASPEN PLUS$^\circledR$ and several standalone software packages for data reconciliation such as VALI, DATACON$^\circledR$, etc., are also available and deployed in chemical and mineral process industries. The main benefit derived from applying DR are accurate estimates of all process variables and parameters which satisfy the process constraints such as material and energy balances.  The derived estimates are typically used in retrofitting, optimization and control applications.  In order to apply DR, the following information is required.

(i) The constraints that have to be obeyed by the process variables and parameters must be defined.  These constraints are usually derived from first principles model using process knowledge, and consist of material and energy conservation equations including property correlations, and can also include equipment design equations, and thermodynamic constraints.    
 
(ii) The set of process variables that are measured must be specified. Additionally, inaccuracies in these measurements must be specified in terms of  the variances and covariances of errors. This information is usually derived from sensor manuals or from historical data. 

Another multivariate data processing technique that has become very popular in recent years is Principal Component Analysis (PCA) \cite{Jolliffe02}.  This method is primarily used for reducing the dimensionality of data and to denoise them.  It is also used in developing regression models, when there is collinearity in the regressors variables \cite{Davies00}.  In chemical engineering, it has  been used for process monitoring and fault detection,  and diagnosis \cite{Kourti95,Yoon01}.  Generally, PCA has been regarded as a data-driven multivariate statistical technique.  In a recent paper, PCA was interpreted as a model identification technique that discovers the linear relationships between process variables \cite{NarasimhanS08}.  This interpretation of PCA is not well known, although other authors have previously alluded to it.  

The purpose of this article is to establish the close connection between PCA and DR. Specifically, it is shown that PCA is a technique that discovers the underlying linear relationships between process variables while simultaneously reconciling the measurements with respect to the identified model.  Exploring this connection further, it is shown that Iterative PCA is a method which simultaneously extracts the linear process model, error-covariance matrix and reconciles the measurements  \cite{NarasimhanS08}.  Several benefits accrue from this interpretation:

(i) It shows that data reconciliation can be applied to a process purely using measured data, even if it is difficult to obtain a model and measurement error variances using \textit{a priori} knowledge. It thus expands the applicability of data reconciliation and related techniques. 

(ii) PCA and IPCA can be used as techniques for obtaining a process model and measurement error-covariance matrix from data.  Since these are the two essential information required to apply DR, it is now possible to apply the rigorous and well developed companion technique such as gross error detection (GED) for fault diagnosis.  This will eliminate the difficulties and deficiencies present in the current approach of using PCA for fault diagnosis.

Additional useful results presented in this paper include the interpretation of the process model obtained using PCA, when only a subset of the process variables are measured.  Modification of the PCA and IPCA techniques to incorporate partial knowledge of some of the process constraints is also proposed.  The impact of incorrectly estimating the model order (the actual number of linear constraints) on the reconciled estimates is also discussed, leading to a recommendation for practical application of PCA and combining it with tools of DR and GED. 

The paper is organized as follows. Sections~\ref{sec:DR} and ~\ref{sec:PCA} introduce the background on DR and PCA, respectively.  Model identification and data reconciliation using PCA for the case of known error-covariance matrix is described in Section~\ref{sec:ModelID}. For unknown error-covariances case, Section~\ref{sec:IPCA} describes a procedure for simultaneous model identification, estimation of error-covariances, and data reconciliation using IPCA. Section~\ref{sec:extensions} extends PCA (IPCA) to partially measured systems, and known constraint matrix. Further, it discusses selection criteria of model order when the model order is not known.  Section~\ref{sec:conclusion} concludes the paper. The developed concepts are illustrated via a simulated flow process.
\section{Basics of Data Reconciliation}\label{sec:DR}
In this section, the application of DR to linear steady--state processes is discussed, including the case when a subset of the process variables is measured (also known as partially measured systems).

\subsection{Linear steady--state processes} \label{sec:LinearDR}
The objective of data reconciliation is to obtain better estimates of process measurements by reducing the effect of random errors in measurements.  For this purpose, the relationships between different variables as defined by process constraints are exploited. We restrict our attention to linearly constrained processes which are operating under steady state.  An example of such a process is a water distribution network, or a steam distribution network with flows of different streams being measured. We first describe the data reconciliation methodology for the case when the flows of all streams are measured.

Let $\xbf(j) \in \Rbb^n$  be an $n$-dimensional vector of  the true values of the $n$ process variables  corresponding to a  steady-state operating  point for each sample $j$. The samples $\xbf(j),\, j=1,2,\ldots,N$ can be drawn from the same steady state or from different steady states.  These variables are related by the following linear relationships:
\be{\Abf\xbf(j)=\zeros_{m \times 1} \label{Eq:constrainform}}
where $\Abf$ is an $(m \times n)$-dimensional matrix, and $\zeros$ is an $m$-dimensional vector with elements being zero.   In data reconciliation, $\Abf$ is labelled as a ``{\it constraint matrix}". Note that the rows of $\Abf$ span an $m$-dimensional subspace of $\Rbb^n$, while $\xbf(j)$ lies in an $(n-m)$-dimensional subspace (orthogonal to the row space of $\Abf$) of $\Rbb^n$. 
Let $\ybf(j)\in \Rbb^n$ be the measurements of the $n$ variables. The measurements are usually corrupted by random errors. Hence, the measurement model can be written as follows:
\be{\ybf(j)=\xbf(j) +\epsilonbf(j), \label{measurementmodel}}
where $\epsilonbf(j)$ is an $n$-dimensional random error vector at sampling instant  $j$.  The following assumptions are made about the random errors:
\bea{(i)& &\ \  \epsilonbf(j)\sim \mathcal{N}(\zeros,\Sigmabf_\epsilon) \nonumber \\
(ii)& & \ \  \Exp{\epsilonbf(j)\epsilonbf(k)^{\tr}}=\zeros,\, \forall \ j\neq k  \nonumber\\
(iii)& &\ \ \Exp{\xbf(j)\epsilonbf(j)^{\tr}}=\zeros \label{eq:errorprops}}
where $\Exp{\cdot}$ denotes the expectation operator. 
If the error variance-covariance matrix $\Sigmabf_\epsilon$ is known, then the reconciled estimates for $\xbf(j)$ (denoted as $\hat \xbf(j)$) can be obtained by minimizing the following objective function:
\bea{\min_{\xbf(j)}\,(\ybf(j) - \xbf(j))^{\tr}\Sigmabf_{\epsilon}^{-1}(\ybf(j) -\xbf(j)) \nonumber\\
s.t.\,\, 
\Abf\xbf(j) =\zeros. \label{eq:vanilladatareconcilation}}
The reconciled values of the variables are given by:
\bea{\hat \xbf(j)=\ybf(j) - \Sigmabf_\epsilon\Abf^{\tr}(\Abf\Sigmabf_\epsilon\Abf^{\tr})^{-1}\Abf\ybf(j)=\Wbf\ybf(j) \label{Eq:recondata},} 
where $\Wbf=\Ibf - \Sigmabf_\epsilon\Abf^{\tr}(\Abf\Sigmabf_\epsilon\Abf^{\tr})^{-1}\Abf$.
Under the assumptions made regarding the measurements errors, it can be shown that the reconciled estimates obtained using the above formulation are maximum likelihood estimates.  It can also be verified that the estimates $\hat{\xbf}(j)$ satisfy the imposed constraints and are normally distributed with mean, $\xbf(j)$, and covariance, $\Wbf\Sigmabf_\epsilon\Wbf^{\tr}$.

\begin{figure}[h!]
\begin{center}

\includegraphics[scale=0.5]{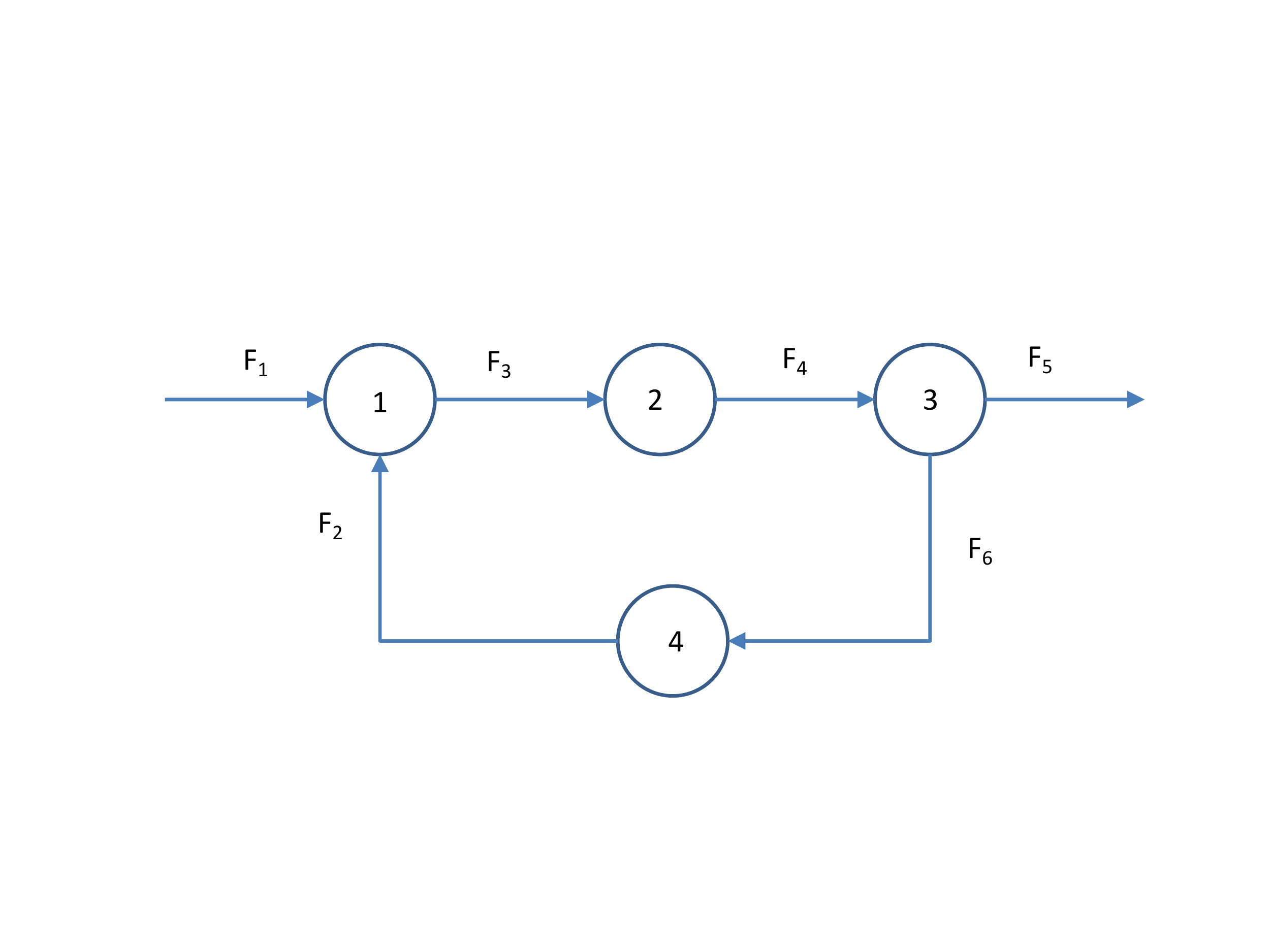}
\caption{Schematic of a flow process}\label{Fig:flow}
\end{center}
\end{figure}

If all the measured samples are drawn from the same steady state operating point, then DR can be applied to the average of the measured samples. However, if the samples are from different steady states, then DR is applied to each sample independently.   For ease of comparison with PCA, we consider a set of $N$ samples (which could correspond to different steady state operating periods) to which DR is applied.  The set of $N$ samples is arranged in the form of an $(n \times N)-$dimensional data matrix, $\Ybf$ as
\be{\Ybf=[\ybf(1),\ybf(2),\ldots,\ybf(N)]= \Xbf + \Ebf,  \label{eq:datamatrix}}  
where $\Xbf$ and $\Ebf$ are  $(n \times N)-$dimensional matrices of the true values and the errors, respectively. The matrix $\hat\Xbf$ of reconciled estimates for the $N$ samples is given by
\be{\hat\Xbf =\Wbf\Ybf. \label{eq:reconcilematrix}} 
The following example illustrates DR on the flow process shown in Figure~\ref{Fig:flow}.\\

\textit{Example 1}\\
  The flow process consists of six flow variables and four balance equations, i.e., $n=6$ and $m=4$.  The flow balance equations for this process can be written as follows:
\bea{\Abf\Fbf&=&\zeros,\,\,\, \label{Eq:example}\text{with}\\
\Abf&=&\bMat{1 & 1 & -1 &  0 & 0 &0\\0& 0& 1& -1& 0& 0\\0& 0& 0& 1& -1 &-1 \\0&-1&0&0&0&1}, \label{eq:trueconstraintmat} \\
\Fbf&=&\bMat{F_1 &F_2& F_3& F_4& F_5 & F_6}^{\tr}. }
In order to demonstrate the utility of DR in reducing noise in measurements, we assume that flows of all six streams are measured and noisy measurements are simulated as follows.  First noise-free values (true flow rates) that satisfy the constraints are generated, followed by addition of noise to generate the noisy measurements. 
\begin{enumerate}
\item The flow variables $F_1$ and $F_2$ are chosen as independent variables and base values are specified for these variables. For simulating data corresponding to different steady states, normally distributed random fluctuations are added to the base values of flow variables $F_1$ and $F_2$. 
\item The true values of the dependent flow variables, $F_3$, $F_4$,  $F_5$, and $F_6$ are computed using four flow balance equations in \eqrefn{Eq:example} for all steady states.
\item Noisy measurements are simulated by adding normally distributed random errors to the true values corresponding to different steady states.
\end{enumerate}
The base values, standard deviations of fluctuations for generating different steady states, and the standard deviations of measurement errors are given in Table~\ref{Table:Simulation}.

\begin{table*}[htdp]
\caption{Base values, standard deviation of fluctuation (SDF), and standard deviations of error (SDE)  for the flow data. }
\begin{center}

\begin{tabular}{llll} \hline
Variable& Base values& SDF& SDE   \\ \hline \hline
$F_1$& 10& 1& 0.1\\\hline
$F_2$& 10& 2& 0.08\\\hline
$F_3$& \eqrefn{Eq:example}& & 0.15\\\hline
$F_4$& \eqrefn{Eq:example}&  & 0.2\\\hline
$F_5$& \eqrefn{Eq:example}&&  0.18\\\hline
$F_6$& \eqrefn{Eq:example}&&  0.1\\\hline
\end{tabular} \label{Table:Simulation}
\end{center}

\end{table*}

A sample of 1000 measurements are simulated and DR is applied to obtain the reconciled estimates corresponding to each sample. The root mean square error (RMSE) values between reconciled and true values over all the 1000 samples for each flow variable is computed and reported in Table~\ref{Table:DRresults}.  The RMSE  in the measurements of variables are also reported in the table for a comparison. The RMSE values in the estimates of all variables are less than the RMSE values in the corresponding measurement, clearly indicating that reconciled estimates of variables are more accurate than the measurements. 
  
\begin{table}[htdp]
\caption{Root mean square errors in measurements and reconciled estimates of flow variables for Example 1}
\begin{center}

\begin{tabular}{lll} \hline
Variable & RMSE before DR & RMSE after DR  \\ \hline \hline
$F_1$&3.2177 &2.3318\\\hline
$F_2$& 2.5750 & 1.6669\\\hline
$F_3$& 4.8292 & 2.5180\\\hline
$F_4$& 6.5358& 2.5180\\\hline
$F_5$& 5.8500& 2.3318\\\hline
$F_6$& 2.7261& 1.6669\\\hline
\end{tabular} \label{Table:DRresults}
\end{center}

\end{table}
\subsection{Partially measured systems} \label{sec:DRpart}
For reasons of cost and feasibility, in most processes only a subset of variables is measured. We refer to such a system as a partially measured system.  Application of DR to a partially measured system provides reconciled values of  measured variables and estimates for  unmeasured variables.  DR also provides diagnostics on which of the measured variables can be reconciled (known as redundant measured variables) and which of the unmeasured variables can be uniquely estimated (also known as observable unmeasured variables). The details of  the redundant and observable variables concepts can be found in \cite{MahR90}.  The method for applying DR to a partially measured system is described below. 

The constraints for a partially measured system, \eqrefn{Eq:constrainform}, can be rewritten in terms of the measured variables (labelled as $\xbf_k$) and the unmeasured variables (labelled as $\xbf_u$) as:
\be{\Abf_k\xbf_k +\Abf_u\xbf_u=\zeros, \label{eq:measureunmeasure}}
where $\Abf_k$ and $\Abf_u$ are the partitions of $\Abf$ matrix corresponding to $\xbf_k$ and $\xbf_u$, respectively. Then, by constructing a projection matrix  $\Pbf$ such that $\Pbf\Abf_u=\zeros$, and multiplying \eqrefn{eq:measureunmeasure}, we can get a reduced set of constraints involving only the measured variables as \cite{Crowe83}:
\be{\Pbf\Abf_k\xbf_k=\zeros. \label{eq:projectedmeasure}}
Then, the reduced data reconciliation problem is to find the minimum of the objective defined in  \eqrefn{eq:vanilladatareconcilation} subject to the constraints defined  by \eqrefn{eq:projectedmeasure}. The reconciled values $\hat\xbf_k(j)$ obtained from the measurement $\ybf_k(j)$ are given as follows:
\be{\hat \xbf_k(j)=\ybf_k(j) - \Sigmabf_\epsilon(\Pbf\Abf_k)^{\tr}(\Pbf\Abf_k\Sigmabf_\epsilon(\Pbf\Abf_k)^{\tr})^{-1}(\Pbf\Abf_k)\ybf_k(j) \label{Eq:recondatameausresubset} } 
The estimates for  the unmeasured variables can be obtained by substituting for the estimates of the measured variables in \eqrefn{eq:measureunmeasure} and solving these equations.  The conditions under which unique estimates for the unmeasured variables can be obtained and the identification of redundant and observable variables using linear algebraic techniques are more completely described in the book by \cite{NarasimhanJ00}. Alternatively, for flow processes, a  graph-theoretic approach can be used for obtaining the reduced constraint matrix, and for determining the redundant measured variables and the observable unmeasured variables.

\begin{figure}[h!]
\begin{center}

\includegraphics[scale=0.5]{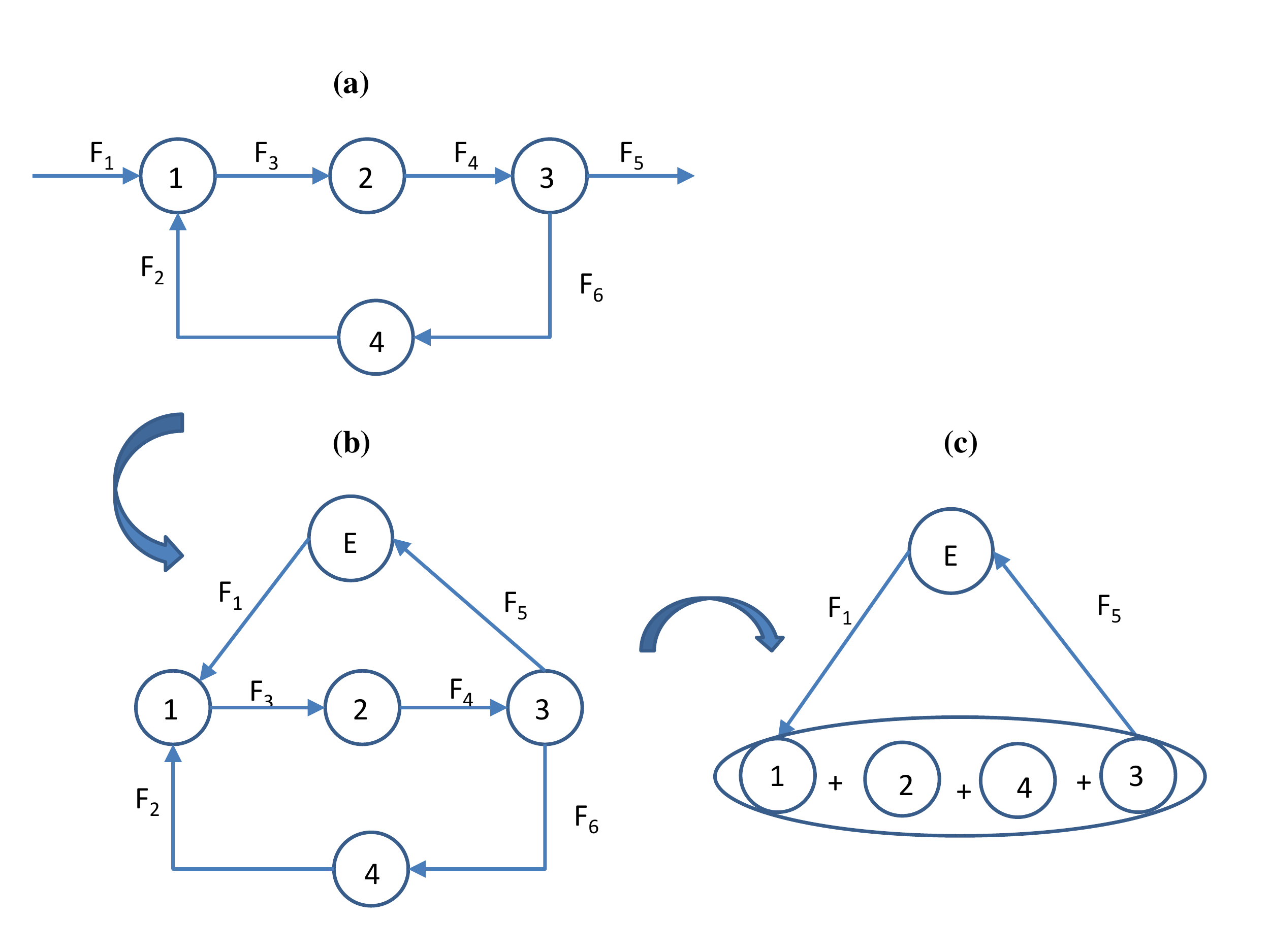}
\caption{Flow network after merging  variables for Example 2}\label{Fig:flowgraph}
\end{center}
\end{figure}

\textit{Example 2} \\
The following  example illustrates the application of DR to a partially measured  flow process. The graph theoretical approach is used for observability and redundancy classification of variables, because it aids in visualization and ease of understanding. 
Consider again the flow process given in Figure~\ref{Fig:flow}, with only the flows of streams 1, 2, and 5 being measured.  In order to apply the graph theoretic procedure, the process is represented as a graph shown in Figure~\ref{Fig:flowgraph}(b), in which an environment node is added to which all process inflows (stream 1) and process outflows (stream 5) are connected. In order to obtain the graph corresponding to the reduced reconciliation problem, we merge nodes which are connected by streams whose flows are not measured.  Thus, merging nodes 1 and 4, 2 and 3, 3 and 4, we get the reduced graph shown in Figure~\ref{Fig:flowgraph}(c). The reduced graph contains only streams 1 and 5 (whose flows are measured) for which one flow balance can be written (either around node 1 or node E).  The reduced DR problem is to obtain reconciled estimates of these two flows subject to the flow balance constraint.  The same reduced reconciliation problem can be obtained using the projection technique described earlier.  The measurements of streams that appear in the reduced graph are redundant, while the measured flow of stream 2 that was eliminated during the merging process is non-redundant.  It can also be deduced that the flows of streams 3, 4, and 6 are observable because the original process graph does not contain any cycle which consists solely of unmeasured flows. 

 The RMSE values in all the variables after DR  are reported in Table~\ref{Example2}. It shows that  the reconciled estimates  of the redundant variables $F_1$ and $F_5$ are more accurate than the measurements even with the partial measurements, while there is no improvement in the non-redundant flow $F_2$. Compared with the RMSE values of estimates reported in Table~\ref{Table:DRresults} for the fully measured case, the estimates of all variables obtained are less accurate due to reduced information available about the process.  

\begin{table*}[htdp]
\caption{Root mean square errors in the reconciled estimates of flow variables for Example 2}
\begin{center}

\begin{tabular}{ll} \hline
Variable &RMSE after DR  \\ \hline \hline
$F_1$&2.7899\\\hline
$F_2$& 2.5750 \\\hline
$F_3$&  3.9033\\\hline
$F_4$& 3.9033\\\hline
$F_5$& 2.7899\\\hline
$F_6$& 2.5750\\\hline
\end{tabular}
\end{center}
\label{Example2}
\end{table*}

\section{Basics of Principal Component Analysis} \label{sec:PCA}
Principal component analysis (PCA) is one of the widely used multivariate statistical techniques. The traditional view of PCA as a technique for dimensionality reduction is explained for the sake of continuity, before we describe how PCA can be used as a technique for identifying a steady state linear model from the data.  

PCA is a linear transformation of a set of variables to a new set of uncorrelated variables called principal components (PCs) \cite{Jolliffe02}. The first PC is a new variable with the highest variance among all linear transformations of the original variables.  The second PC is a new variable orthogonal to the first PC (and hence uncorrelated with it) that has the next highest variance among all linear transformations of the original variables and so on.  For reducing the dimensionality of multivariate data, the PCs with the highest variances are retained while the remaining PCs are discarded. The PCs can be obtained from the eigenvectors of the covariance matrix of the data for the given set of variables.  The eigenvectors of the covariance matrix of the data can in turn be obtained using the singular value decomposition (SVD) of the data matrix as described below.

As defined in the preceding section, let $\Ybf$ be the $(n \times N)-$dimensional data matrix.  Let $\Sbf_{\ybf}$ be the covariance matrix of $\Ybf$ defined by
\be{\Sbf_{\ybf} = \frac{1}{N}\Ybf\Ybf^{\tr}. \label{Eq:covdata}}
The SVD of the scaled data matrix can be written as
\be{svd(\frac{\Ybf}{\sqrt{N}}) =\Ubf_1\Sbf_1\Vbf_1^{\tr} + \Ubf_2\Sbf_2\Vbf_2^{\tr}, \label{Eq:denoising} }
where $\Ubf_1$ are the orthonormal eigenvectors corresponding to the  $p$ largest eigenvalues of $\Sbf_{\ybf}$ while $\Ubf_2$ are the orthonormal eigenvectors corresponding to the remaining $(n-p)$ smallest eigenvalues of $\Sbf_{\ybf}$.  The matrices $\Sbf_1$ and $\Sbf_2$ are diagonal matrices, whose diagonal elements are the square root of the eigenvalues of $\Sbf_{\ybf}$.  It is assumed that the eigenvalues and corresponding eigenvectors are ordered in decreasing order of the magnitudes of the eigenvalues.  It can be proved that the first $p$ PCs are given by $\Ubf_1^{\tr}\ybf(j)$, and the variances of these PCs are the corresponding eigenvalues.  

If the objective is to reduce the dimensionality of the data, then the values of the first $p$ PCs only need to be computed corresponding to each observation $\ybf(j)$ and stored.  Several heuristics have been suggested in the literature to choose the number of PCs $p$ to be retained, such as percentage of total variance in data captured, or the SCREE plot which looks for sharp changes in the eigenvalues.  More details on such heuristics can be found in the book by  \cite{Jolliffe02}. 

Alternatively, PCA can be used as  a denoising technique.  The denoised estimates of the measurements obtained using PCA corresponding to $p$ retained PCs are given by
\be{\hat\Xbf = \sqrt{N}\Ubf_1\Sbf_1\Vbf_1^{\tr}. \label{eq:truncateddata}}

\section{Model Identification and Data Reconciliation using PCA}\label{sec:ModelID}

 The viewpoint that is of importance in this article is that PCA is a method for discovering the underlying linear relationships between variables, while at the same time obtaining the reconciled estimates of variables that satisfy the identified linear relations.  The use of PCA as a method for identifying a linear model is less well known although it has been alluded to by different authors \cite{Gertler99,Huang01,Jolliffe02}.  This viewpoint is explored thoroughly in this work. We first analyze the case when all the process variables are measured and the measurement error-covariance matrix is known.  The generalization to partially measured systems and estimation of error-covariance matrix simultaneously along with the linear constraint model from data will be elucidated in the following sections.  We also make the following additional assumptions regarding the observations.  The reasons for these additional assumptions will be explained later in this section.  
\begin{description}
\item[(A1)] The data are drawn from at least $n$ distinct steady states. 
\item [(A2)] The data used for model building do not contain any outliers or gross errors.
\end{description}
If there are $m$ linear relations between the measured process variables as defined by Eq.~(\ref{Eq:constrainform}), then as stated earlier, the true values of variables lie in an $\Rbb^{n-m}$ dimensional subspace of $\Rbb^n$. Therefore, the denoised estimates of variables are also chosen to lie in an $(n-m)$ dimensional subspace.  If we assume that the number of linear constraints $m$ relating the variables is known \emph{a priori}, then we can apply PCA, choose the number of retained PCs $p$ to be equal to $n-m$, and obtain the estimates $\hat{\Xbf}$ given  by Eq.~(\ref{eq:truncateddata}).  Based on the orthogonality of the PCs, it immediately follows that the eigenvectors corresponding to the smallest $m$ eigenvalues are orthogonal to these estimates, and therefore they provide an estimate of the rows of the constraint matrix $\Abf$ (whose $m$ rows lie in an $R^m$ dimensional subspace orthogonal to the true values of process variables), that is, the estimate of the constraint matrix is given by
\be{\hat{\Abf} = \Ubf_2^{\tr}. \label{Eq:estimatedA} }
While data compression and denoising focuses on the eigenvectors corresponding to the largest eigenvalues that need to be retained, model identification is concerned with the eigenvectors corresponding to the smallest eigenvalues. The main questions to be answered are whether the estimate of the constraint matrix obtained using PCA is optimal, and whether the number of constraints can also be estimated from data, without any prior knowledge. 

The estimates of process variables (that lie in a $p$-dimensional subspace) obtained using Eq.~(\ref{eq:truncateddata}) and the corresponding estimate of the constraint matrix (orthogonal to the estimates) obtained using Eq.~(\ref{Eq:estimatedA}), can be shown to be optimal in the least-squares sense \cite{Rao64}. In other words, given a sample of observations, PCA identifies an optimal $p$-dimensional subspace in which the estimates lie such that the sum squared differences between measured and estimated values is minimum.  If we assume that the errors in measurements obey the assumptions listed in Eq.~(\ref{eq:errorprops}) with the additional condition that the error-covariance matrix is a scale of the identify matrix, i.e.,  $\Sigmabf_\epsilon=\sigma^2\Ibf$, then the following result can be proved
\be{\Sigmabf_{\Ybf|\Xbf, N} = E[\Sbf_{\ybf}| \Xbf, N] = \Mbf_{\xbf} + \sigma^2\Ibf, \label{eq:unityev}} 
where $\Mbf_{\xbf}= \frac{1}{N}\Xbf\Xbf^{\tr}$ is an $(n \times n)-$dimensional matrix, and $\Ibf$ is an $n$-dimensional identity matrix. Note  that the notation $E[\Sbf_{\ybf}| \Xbf, N]$ is used to emphasize that  the expected value of the sample covariance matrix is considered under the assumption that the samples are drawn from the same set of steady states $\Xbf$, and  the only variability in $\Sbf_y$ is due to the measurement errors.  \eqrefn{eq:unityev} leads to the following properties:
\begin{property}
  \label{Property1} The eigenvectors of $\Sigmabf_{\Ybf|\Xbf, N}$ and 
  $\Mbf_{\xbf}$ are identical.
\end{property}
\begin{property} \label{Property2}
The smallest $m$ eigenvalues of $\Sigmabf_{\Ybf|\Xbf, N}$ are all equal to $\sigma^2$.
\end{property}
Assumption (A2) made regarding the observations implies that the observations do not contain any systematic error, and, therefore, the true values corresponding to each sample must satisfy the true process constraints. Furthermore, Assumption (A1) ensures that observations from different steady states are obtained.  If data corresponding to $(n-m)$ linearly independent steady states are obtained, then the rank of the matrix $\Mbf_{\xbf}$ will be $(n-m)$ with the last $m$ eigenvalues being equal to zero.\footnote{Since the number of constraints may not be known \emph{a priori}, it is recommended that data be obtained from as many distinct steady state operating conditions as the number of variables $n$.} The eigenvectors corresponding to the $m$ zero eigenvalues will be orthogonal to the true values and is a basis for the row space of the constraints.  We can  therefore use Property \ref{Property1} to conclude that the eigenvectors corresponding to the $m$ smallest eigenvalues of $\Sigmabf_{\Ybf|\Xbf,N}$ is  a basis for the row space of the constraint matrix.  Furthermore, Property \ref{Property2} clearly provides a way of identifying the number of constraints, without any \emph{a priori} knowledge by examining the eigenvalues of $\Sigmabf_{\Ybf|\Xbf,N}$. The number of constraints (also referred to model-order) is estimated to be equal to the number of small eigenvalues all of which should be equal.  
Thus, if $\Sigmabf_{\Ybf|\Xbf,N}$ is  known, we can derive the number of constraints and the constraint matrix by applying PCA to this matrix. However, we can only obtain an unbiased estimate of $\Sigmabf_{\Ybf|\Xbf,N}$ from the sample covariance matrix $\Sbf_{\ybf}$. It has been proved  that the eigenvectors of $\Sbf_{\ybf}$ are asymptotically unbiased estimates of the eigenvectors of $\Sigmabf_{\Ybf|\Xbf,N}$, if the errors and (hence the measurements) are normally distributed \cite{Jolliffe02}. Using this result, we can therefore conclude that if the errors in all measurements are normally distributed with identical variances (homoscedastic errors) and  are mutually independent, then PCA can be used to derive an asymptotically unbiased estimate of the constraint matrix.  It should be noted that the estimate of the constraint matrix derived using PCA can differ from the true constraint matrix form (we may desire) by a rotation matrix.  In other words,  the estimated and true constraint matrices are related as  
\be{\hat{\Abf} = \textbf{Q}\Abf, \label{eq:rotambiguity}}
where $\textbf{Q}$ is some non-singular matrix. 

It can also be shown that the denoised estimates obtained using PCA given by \eqrefn{eq:truncateddata} are the reconciled values of the data matrix $\Ybf$ corresponding to the constraint matrix $\hat{\Abf}$ identified using PCA, under the assumption that $\Sigmabf_\epsilon=\sigma^2\Ibf$.  By substituting the estimated constraint model given by \eqrefn{Eq:estimatedA} in  \eqrefn{Eq:recondata}, we obtain 
\be{\hat\Xbf = (\Ibf - \Ubf_2(\Ubf_2^{\tr}\Ubf_2)^{-1}\Ubf_2^{\tr})\Ybf. \label{eq:reconest}}
Substituting for the SVD of the data matrix and using the property that the eigenevctors are orthonormal, we get
\be{\hat\Xbf = \sqrt{N}(\Ibf - \Ubf_2\Ubf_2^{\tr})(\Ubf_1\Sbf_1\Vbf_1^{\tr} + \Ubf_2\Sbf_2\Vbf_2^{\tr}). \label{eq:intermediatproof1}}
Expanding the above equation and using the orthonormal property of the eigenvectors we finally get
\be{\hat\Xbf = \sqrt{N}\Ubf_1\Sbf_1\Vbf_1^{\tr}.\label{eq:finalproof1}}

Based on the above analysis, it can be concluded that PCA is a method that simultaneously identifies the constraint model and obtains reconciled estimates with respect to the identified model purely from data under the assumption that measurement errors in different variables are independent and have the same variance. The number of constraints can also be obtained purely from data.  It should, however, be noted that while the estimates obtained using PCA satisfy the identified constraint matrix, they will not satisfy the true process constraints due to inaccuracies in the estimated constraint matrix.  The larger the sample size of the data set, the more accurate will be the estimated constraint matrix, and more closely will the PCA estimates match the reconciled estimates derived by applying DR using the true process constraints.  It may also be verified that the reconciled estimates are invariant to a rotation of the constraint matrix.  Thus, the fact that the constraint matrix estimated using PCA differs from the desired form of the true constraint matrix (derived from first principles) by a rotation, will not have an effect on the reconciled estimates. 
 
\subsection{Error-covariance matrix known}\label{section:errorcovknown}
The results in the preceding subsection were obtained under the assumption that the errors in measurements of different variables have identical variances.  In this subsection, we describe the identification of constraint model using PCA when the measurement errors have different variances and may also, in general, be correlated.  It is however assumed that the error-covariance matrix $\Sigmabf_\epsilon$ is known.  Furthermore, the errors in measurements of different samples are assumed to be independent and identically distributed with the same known covariance matrix.  It may be noted that these assumptions regarding the measurement errors are the same as those made in DR.   

Narasimhan and shah\cite{NarasimhanS08} described an approach for applying PCA for model identification from data when the error-covariance matrix is known.  This approach is based on transforming the data using an appropriate matrix before applying PCA. The measurement errors corrupting the transformed data are independent and identically distributed (i.i.d.), and PCA can be applied to the transformed data matrix. The approach is as follows:
 
The Cholesky decomposition of $\Sigmabf_\epsilon$ is given by
 \be{\Sigmabf_\epsilon=\Lbf\Lbf^{\tr}, \label{Eq:Cholesky}}
where $\Lbf$ is an $(n \times n)-$dimensional  lower triangular matrix.   Then, the noisy data matrix can be transformed as follows:
\be{\Ybf_{s} = \Lbf^{-1}\Ybf = \Lbf^{-1}\Xbf +\Lbf^{-1}\Ebf, \label{eq:transformdata}}
where $\Ebf$ is an $(n \times N)-$dimensional error matrix. Let $\Sbf_{\ybf_s}$ is the covariance matrix of the transformed data defined by
\be{\Sbf_{\ybf_s} = \frac{1}{N}\Ybf_s\Ybf_s^{\tr}}
By taking expectation we can show that
\be{\Sigmabf_{\Ybf_s|\Xbf, N} = E[\Sbf_{\ybf_s}|\Xbf, N] = \Mbf_{\xbf_s} + \Ibf, \label{eq:unityevscaled}} 
where $\Mbf_{\xbf_s} = \frac{1}{N} \Lbf^{-1}\Xbf\Xbf^{\tr}\Lbf^{-\tr}$. 
Eq.~(\ref{eq:unityevscaled}) is similar \eqrefn{eq:unityev} and Properties 1 and 2 hold for the matrix $\Sigmabf_{\Ybf_s|\Xbf, N}$.  Therefore, PCA can be applied to the transformed data in order obtain the constraint model and reconciled estimates corresponding to the transformed data.  The reconciled estimates and constraint model corresponding to the original data can be derived using an inverse transformation as described below.
Let the SVD of $\Ybf_s$ be decomposed corresponding to the first largest $n-m$ singular values and remaining $m$ singular values as
\be{\Ybf_s = \underbrace{\sqrt{N}\Ubf_{1s}\Sbf_{1s}\Vbf_{1s}^{\tr}}_{\hat{\Xbf}_s}+ \sqrt{N}\Ubf_{2s}\Sbf_{2s}\Vbf_{2s}^{\tr}. \label{eq:scaleddatasvd} }
The first part on the right-hand side of above equation corresponds to the reconciled estimates of the transformed data, while the second part contains the information about the constraints which relate the transformed variables.  The reconciled estimates and the constraint matrix corresponding to the original data are given by 
 \bea{\hat\Xbf&=&\Lbf\hat{\Xbf}_s \label{eq:unscaledest}\\
 \hat{\Abf}&=&\Ubf_{2s}^{\tr}\Lbf^{-1}. \label{eq:unscaledmatrix}}
Since the smallest $m$ eigenvalues of $\Sbf_{\ybf_s}$ matrix are all equal to $1$, it provides a systematic method to determine the constraint model order.  Due to finite sample sizes and numerical inaccuracies, the equality of smallest $m$ eigenvalues can be numerically checked by testing whether the average of the  smallest $m$ eigenvalues  approximately equals unity. 

\textit{Example 3} \\
The flow process of Example 1 is again considered along with the simulated data set of 1000 samples. We assume that the error-covariance matrix (used to simulate the measurements) is known, but neither the number of constraints nor the constraint matrix is assumed to be known.  It may be noted that the error-covariance matrix is diagonal, with the diagonal elements being the error variances as given in Table~\ref{Table:Simulation}.  The data matrix is transformed as defined by \eqrefn{eq:transformdata} and its SVD obtained.  The six ordered singular values obtained are 263.16, 21.83, 1.05, 1.04, 1.01, and 0.98.  Examination of the singular values clearly indicates that there are four constraints, because the smallest four singular values are all nearly equal.  The denoised estimates of the transformed data are obtained using the first two PCs, and the reconciled estimates of the original data are obtained using \eqrefn{eq:unscaledest}.  The RMSE in reconciled estimates of each variable is computed and reported in Table~\ref{Table:PCAresults}.  Comparing the RMSE of estimates obtained using PCA with those reported in Table~\ref{Table:DRresults} which are obtained by applying DR using the known process model, it is clear that they are nearly equal. Although the PCA estimates do not exactly satisfy the true flow balances given by \eqrefn{Eq:example}, the maximum constraint residual is of the order of $10^{-12}$.  These indirectly indicate that the constraint model obtained using PCA is a good estimate of the true process constraints.

\begin{table}[htdp]
\caption{Root mean square errors in the reconciled estimates of flow variables using PCA for Example 3}
\begin{center}

\begin{tabular}{ll} \hline
Variable & RMSE using PCA \\ \hline \hline
$F_1$& 2.3383\\\hline
$F_2$& 1.6758 \\\hline
$F_3$&  2.5225\\\hline
$F_4$&  2.5288\\\hline
$F_5$&  2.3290\\\hline
$F_6$&  1.6682\\\hline
\end{tabular} \label{Table:PCAresults}

\end{center}
\end{table}
\subsection{Comparison of PCA estimated constraint model and true process constraint model}
In Example 3, the eigenvectors of the transformed data covariance matrix corresponding to the smallest four eigenvalues are used to derive an estimate of the process constraint matrix using \eqrefn{eq:unscaledmatrix}, and is given below
\be{\hat{\Abf} = \bMat{ -2.4832 &   0.2270 &  -2.4441  &  4.3450  &  0.5926 &  -2.1390\\
    1.5738  & -5.0224 &   1.4541  &  1.4337  & -4.4687 &   2.1376\\
    5.3860  &  6.5976 &   -2.9594  & -0.3256&   -2.0764 &  -3.3396\\
    2.9833 &  -3.4863 &  -3.8879 &  -0.1440 &   1.0598   & 7.5021}. \label{eq:estconstraintmat}}
The elements of the estimated constraint matrix do not seem to correspond with those of the constraint matrix derived from the first principles given by \eqrefn{eq:trueconstraintmat}. However, it should be noted that an element by element comparison between the estimated and true constraint matrices cannot be done, because the estimated constraint matrix may differ from the first principles model by a rotation matrix (see \eqrefn{eq:rotambiguity}). Thus, only the row spaces of  the estimated and true process constraint matrices can be compared.  Two criteria were proposed by \cite{NarasimhanS08} for making such a comparison (i) the subspace angle between the row subspaces of the estimated and true constraint matrices, and (ii) the sum of orthogonal distances of the row vectors of the estimated constraint matrix from the subspace defined by the rows of the true constraint matrix. This metric (denoted by $`\alpha$') is computed as follows:
          \bea{\alpha&=&\Sigma_i \alpha_i, \hspace{1cm}\text{with}\\
          \alpha_i&=&\|\hat\Abf_{i.}-\hat\Abf_{i.}\Abf^{\tr}(\Abf\Abf^{\tr})^{-1}\Abf\|,\label{eq:alpha}}          
where $\hat{\Abf}_{i.}$ are the rows of the estimated constraint matrix.  Hence,  $\alpha$ value near to zero indicates that $\hat\Abf$ is a good approximation of  $\Abf$.

For Example 3, the subspace angle between the row spaces of estimated and true constraint matrices is 0.2377 degrees and $\alpha$ = 0.0555.  These values suggest that a good estimate of the constraint matrix has been obtained using PCA.  Our experience with larger examples indicates that subspace angle is not a good criteria for comparison, because it does not clearly indicate the quality of the estimated constraint matrix, especially as the number of constraints increases.  

A  practically useful method for comparing the estimated and true constraint matrix is proposed in this work.  For this purpose, the process variables are partitioned into a set of dependent variables $\xbf_D$ and a set of independent variables $\xbf_I$, based on process knowledge.  The number of dependent variables should be chosen equal to the number of constraints.  The constraints given by \eqrefn{Eq:constrainform} can be rewritten as
\be{\Abf_D\xbf_D + \Abf_I\xbf_I = 0, \label{eq:constdepform}}     
where $\Abf_D$ and $\Abf_I$ are the sub-matrices of $\Abf$ corresponding to dependent and independent variables, respectively.  From \eqrefn{eq:constdepform}, the regression matrix relating the dependent variables to the independent variables can be obtained as
\be{\xbf_D = -(\Abf_D)^{-1}\Abf_I\xbf_I = \Rbf\xbf_I.} 
In a similar manner, the estimated regression matrix can be obtained from the estimated constraint matrix as 
\be{\hat{\Rbf} = -\hat{\Abf}_D^{-1}\hat{\Abf}_I} 
An element by element comparison can be made between $\Rbf$ and $\hat{\Rbf}$. 

For Example 3,  if the flow variables $F_3$ to $F_6$ are chosen as the dependent variables, then the true regression matrix and the estimated regression matrix derived from the estimated constraint matrix are given by
\bea{
\Rbf &=&\bMat{1 & 1\\1& 1\\1 & 0\\0& 1}\\
\hat{\Rbf}&=& \bMat{1.0057 & 0.9936\\1.0013& 0.9988\\1.0013 & -0.0022\\0.0013& 0.9991}.
}

An element by element comparison between the two regression matrices shows that the maximum absolute difference between the elements is  0.0064.  This again clearly shows that an accurate estimate of the constraint matrix is obtained using PCA.  

\section{Model Identification, Error Variance Estimation and Data Reconciliation using IPCA} \label{sec:IPCA}
In Section~\ref{section:errorcovknown}, it was established that PCA is a technique for obtaining the steady-state constraint model and reconciled estimates from measured data when the error-covariance matrix is known. In practice, the error variances and covariances are not easily available and may also change with time. If replicate measurements are available corresponding to one or more steady states, then the error-covariance matrix can be directly estimated from the data, provided the measurements corresponding to each steady state operating period are clearly identified. However, identification of steady operating periods from data is itself a challenging problem. Given these practical difficulties, the question is whether it is possible to simultaneously estimate the error-covariance matrix, the process constraint model, and the reconciled estimates purely from data, without the need for replicate measurements corresponding to a steady state. Surprisingly, it is possible to extract all this information from data.  A method for this purpose known as iterative PCA (IPCA) was recently proposed by  \cite{NarasimhanS08}. A brief description of this method follows.

In Section \ref{section:errorcovknown}, it was shown how PCA can be used to obtain an estimate of the constraint matrix and reconciled estimates if the error-covariance matrix is given.  In IPCA, this approach is iteratively combined with an algorithm for estimating the error-covariance matrix from data, given the constraint matrix, by solving the following optimization problem.
\begin{equation} \label{eq:varestmle}
\min_{\Sigmabf_{\epsilon}}
N\log|\hat{\Abf}\Sigmabf_{\epsilon}\hat{\Abf}^{\tr}| +
\sum_{k=1}^{N}[\rbf^{\tr}(k)(\hat{\Abf}\Sigmabf_{\epsilon}\hat{\Abf}^{\tr})^{-1}\rbf(k)]
\end{equation}
Under the assumptions made regarding the measurement errors \eqrefn{eq:errorprops}, the objective function used for the estimating the error-covariance matrix is identical to maximizing the likelihood function of the constraints residuals $\rbf(k)$ defined by
\be{\rbf(k) = \hat{\Abf}\ybf(k) \label{eq:constres}}
An updated estimate of the constraint matrix can be obtained using the estimated error-covariance matrix as described in Section \ref{section:errorcovknown}, and the procedure repeated until the estimates of the error variance matrix (and constraint matrix) converges.  At convergence, the smallest $m$ eigenvalues should be all equal to unity.  Thus, the convergence test can be applied to check if the average of the smallest $m$ eigenvalues is close to unity.    

The method described in this section for estimating the measurement error-covariance matrix from constraints residuals is known as an indirect method in the area of data reconciliation.  Using indirect methods, the maximum number of elements of the symmetric error-covariance matrix that can be estimated is $m(m+1/2)$.  Almasy et al. \cite{Almasy84} were the first to propose an indirect method, and obtained a solution which minimizes the sum of the off-diagonal elements of the error-covariance matrix.  Later, Keller et al\cite{Keller92} obtained an analytical solution for the least squares solution for the elements of error-covariance matrix. A robust indirect approach for estimating the error variances was also proposed by \cite{ChenBR97}.  The approach described in this section is a maximum likelihood estimation procedure. Besides, a significant difference between the procedure described here and other indirect approaches proposed earlier in the field of DR, is that the constraint matrix is assumed to be known in earlier approaches, whereas the procedure described here is combined with PCA for simultaneous estimation of the constraint matrix and error-covariance matrix.  The reconciled estimates of variables corresponding to these estimated constraint and error-covariance matrix are also simultaneously obtained using \eqrefn{eq:unscaledest}.  

\textit{Example 4} \\
The flow process of Example 1 is  considered to demonstrate IPCA. It is assumed that the error-covariance matrix is diagonal, but its elements are unknown. We need to estimate six elements of $\Sigmabf_\epsilon$, along with the constraint matrix and reconciled estimates.  The number of constraints for this example is four, and it is possible to estimate a maximum of $4(4+1)/2=10$ elements of the error-covariance matrix.  Since we are estimating only $6$ diagonal elements of  $\Sigmabf_\epsilon$, the problem is identifiable.  IPCA is applied to the data matrix and the estimated standard deviations of errors are given in Table~\ref{Tab:Example4}.  The results show that IPCA provides accurate estimates of error variances. Further, the six ordered singular values obtained are  258.47, 21.53, 1.02, 1, 0.99, and 0.99, which clearly indicates that there are four constraints. The reconciled values and the constraint matrix can be estimated using Eqs.~\eqref{eq:unscaledest} and \eqref{eq:unscaledmatrix}. The subspace angle between the row spaces of estimated and true constraint matrices is 0.2373 degrees and $\alpha$ = 0.0509. The maximum constraint residual is of the order of $10^{-12}$. The estimated regression matrix $(\hat{\Rbf}_{ipca})$ computed by considering the flow variables $F_3$ to $F_6$ as dependent variables is as follows:
  \bea{
   \hat{\Rbf}_{ipca}&=& \bMat{1.0056 & 0.9937\\1.0012& 0.9990\\1.0012 & -0.0022\\0.0012& 0.9993}.
  }
  The maximum absolute difference between the elements of $\Rbf$ and $\hat{\Rbf}_{ipca}$ is  0.0063. The comparison of the estimated constraint matrix using the subspace angle ($\alpha$), and $\hat{\Rbf}_{ipca}$ shows that IPCA provides an accurate estimate of the constraint matrix without knowledge of the error-covariance matrix.  Further, the RMSE values in the reconciled estimates of each variable are computed and reported in Table~\ref{Tab:Example4}. The RMSE values  show that the reconciled estimates are  comparable to the one obtained by applying DR using the known constraint model. The simulation results shows that IPCA is a reliable data-driven method for obtaining   the reconciled values, a constraint matrix, and estimates of standard deviations from the data without any \emph{a priori} knowledge.
 \begin{table}[htdp]
 \caption{Standard deviation (SD) and root mean square errors in the reconciled estimates of flow variables using IPCA (Example 4)}
 \begin{center}
 
 \begin{tabular}{lll} \hline
 Variable & Estimated SD& RMSE using IPCA  \\ \hline \hline
 $F_1$&0.1006 & 2.3439\\\hline
 $F_2$& 0.0848 & 1.6753 \\\hline
 $F_3$&  0.1507& 2.5246\\\hline
 $F_4$&  0.2128& 2.5309\\\hline
 $F_5$&  0.1873& 2.3344\\\hline
 $F_6$&  0.0866& 1.6689\\\hline
 \end{tabular}  \label{Tab:Example4}
 \end{center}

 \end{table}

\section{Extensions to PCA based Data Reconciliation} \label{sec:extensions}
\subsection{Partially measured systems}
In the preceding sections, the close link between DR and  PCA was established.  Further, we have demonstrated that it is possible to identify the constraint model and error-covariance matrix, if required, from data using PCA (or IPCA). The estimated model and error-covariance matrix can be subsequently used for obtaining more accurate estimates of variables, by reconciling the measurements with respect to the identified model.  In deriving these results, it was assumed that all process variables are measured, which is not valid in general.  The connection between PCA and DR applied to a partially measured system is elucidated in this section.  

We consider a process defined in Section~\ref{sec:DRpart}. Let $n_p$ be the number of measured variables out of $n$ process variables, and let $\ybf_p(j)$ be measurements of $n_p$ variables at the $j$th sample, where $j=1,\ldots, N$. The corresponding $(n_p \times N)-$dimensional measurement matrix $\Ybf_p$ can be obtained by collating $\ybf_p(j)$ for all $N$ samples.  Then, the objective is to identify a linear model relating the measured variables and obtain reconciled estimates of these variables.

If the measurement error-covariance matrix is known, then we can apply PCA as described in Section \ref{section:errorcovknown} to the data matrix $\Ybf_p$ and estimate the number of constraints, linear constraint matrix and reconciled values of the measured variables.  On the other hand, if the measurement error-covariance matrix is unknown, then we can apply IPCA as described in Section \ref{sec:IPCA} to simultaneously estimate the error-covariance matrix, constraint matrix and reconciled values.  The key question is how these estimates derived using PCA or IPCA are related to the estimates obtained by applying DR to the partially measured system described in Section \ref{sec:DRpart}, which uses knowledge of the true process constraints and error-covariance matrix.  It may be noted that by applying PCA or IPCA to the data matrix $\Ybf_p$, it is possible to identify the linear constraints relating only the measured variables. The constraints relating only the measured variables can also be derived using a projection matrix on the known true process constraints as described in Section \ref{sec:DRpart}.  The reduced data reconciliation problem obtained after projecting out the unmeasured variables is identical to the DR problem for the completely measured process, with the only difference being that the constraint matrix relating the measured variables is the reduced balance matrix $\Pbf\Abf_k$.  Therefore, using similar arguments as described in Sections \ref{section:errorcovknown} and \ref{sec:IPCA}, we can derive the following result  
\be{\Ubf_{2p} = \Qbf\Pbf\Abf_x, \label{eq:estApart}}
where $\Ubf_{2p}$ are the eigenvectors of the covariance matrix of $\Ybf_p$ corresponding to the smallest $m_p$ eigenvalues, $\Qbf$ is an arbitrary non-singular rotation matrix and $\Pbf\Abf_x$ is the reduced constraint matrix.  The number of constraints in the reduced reconciliation problem $m_p$ can be shown to be equal to $(m-t)$ where $t$ is the rank of $\Abf_u$ \cite{NarasimhanJ00}. The number of constraints $m_p$ can also be obtained from the data because the smallest $m_p$ eigenvalues should all be equal.  The reconciled estimates corresponding to the identified model are given as before by
\be{\hat{\Xbf}_p = \Ubf_{1p}\Sbf_{1p}\Vbf_{1p}. \label{eq:estXpart} }

While the above discussion brings out the relationship between PCA and DR as applied to a partially measured system, it should be pointed out that because there is no information regarding the unmeasured variables, PCA (or IPCA) can be used to estimate only the reduced constraint matrix that relate the measured variables.  The true constraint matrix that relates both the measured and unmeasured variables cannot obviously be estimated from measured data.  Furthermore, using the first principles model, it is also possible to classify unmeasured variables as observable or unobservable based on the given measurement structure, which cannot be derived using the data driven approach. The classification of measured variables as redundant or non-redundant can, however, be performed using the data driven approach, by examining the columns of the estimated reduced constraint matrix.  If all the elements of a column of $\hat{\Abf}_p  $ are close to zero, then the corresponding measured variable does not participate in any of the reduced constraints and is therefore a non-redundant measured variable.  The following example illustrates the application of PCA to a partially measured flow process.
 
\textit{Example 5} \\  The flow process of Example 2 is considered here. We assume that the error-covariances for these variables are known. Then, PCA approach described in Section~\ref{section:errorcovknown} can be applied to the available flow measurements. The three ordered singular values obtained are 169.84, 18.96, and 1.03. This indicates that there is one constraint among the measured variables. The estimated constraint row is $[ -4.8612,\,\,    0.0107,\,\,    4.8549]$. The estimated constraint is a scale of the projected constraint obtained from first principles obtained in Example 2. Note that the coefficient corresponding to the flow $F_2$ is almost zero, which indicates that $F_2$ is a non-redundant variable and its measurement cannot be reconciled. Comparing the RMSE values of estimates obtained using PCA in Table~\ref{Table:Example5} with the RMSE values of measurements reported in Table~\ref{Table:DRresults}, it is clear that  PCA reconciles the flow variables $F_1$ and $F_5$, but not $F_2$. The maximum constraint residual is of the order of $10^{-13}$ which indicates that the identified model is close to the true one.
\begin{table}[h]
\begin{center}
 \caption{Root mean square errors in the reconciled estimates of flow variables for the partially measured system in Example 5}
 \begin{tabular}{cc} \hline
 Variable & RMSE using PCA with the known SD \\ \hline \hline
 $F_1$& 2.7951\\\hline
 $F_2$& 2.5749 \\\hline
 $F_5$&   2.7873\\\hline
\end{tabular}  \label{Table:Example5}
 \end{center}

 \end{table}
\subsection{Partially known constraint matrix}
In Section \ref{sec:ModelID}, it was shown how PCA can be used for simultaneous model identification and data reconciliation when no knowledge of the constraint matrix is available. In some cases, a subset of constraint matrix (or linear relationships) may be known. In such a situation, it is required to identify only the remaining constraints that relate the measured variables. In this subsection, we extend the approach of PCA and IPCA so that the partial knowledge available about the process constraints can be fully exploited. 

Let $\Abf_g$ be the $(m_g \times n)-$dimensional known constraint matrix. For simplicity we assume that the error-covariance matrix is known and is an identity matrix.  The procedure described in this section can be generalized using the approach described in Sections \ref{sec:ModelID} and \ref{sec:IPCA}, if the error-covariance matrix is not an identity matrix, or if it is unknown. We are interested in identifying from the measured data only the linear constraints other than those that are specified.  We ensure that none of the given constraints or any linear combination of these can be identified by first determining the component of the measured data that is orthogonal to the given constraints.  The orthogonal component of the data is given by
\bea{\Ybf_{proj}=[\Ibf - \Abf^{\tr}_g(\Abf_g\Abf^{\tr}_g)^{-1}\Abf_g]\Ybf.  \label{Eq:projection}}
PCA is applied using $\Ybf_{proj}$ to estimate the remaining $(m-m_g)$ constraints. In order to identify the remaining constraints from the SVD of the projected data matrix, it should be first noted that the projected data matrix satisfies the given constraints and will therefore have a rank equal to $(n-m_g)$.  The smallest $m_g$ singular values of the data matrix will be equal to zero and the corresponding left singular vectors (eigenvectors of the covariance matrix of the projected data matrix) will be exact linear combinations of the given constraints.  Out of the non-zero singular values, we consider the smallest $(m-m_g)$ singular values (all of which should be almost equal).  The SVD of the projected data matrix is therefore partitioned corresponding to the largest $(n-m)$, next largest $(m-m_g)$, and remaining smallest $m_g$ singular values as  
\be{svd(\Ybf_{proj}) = \Ubf_{proj,1}\Sbf_{proj,1}\Vbf_{proj,1}^{\tr} + \Ubf_{proj,2}\Sbf_{proj,2}\Vbf_{proj,2}^{\tr} + \Ubf_{proj,3}\Sbf_{proj,3}\Vbf_{proj,3}^{\tr}}
The transpose of $\Ubf_{proj,2}$ is an estimate of the remaining $(m-m_g)$ constraints. The complete constraint matrix can be constructed as
\be{\hat{\Abf} = \left[ \begin{array}{l}  \Ubf_{proj,2}^{\tr} \\ \Abf_g \end{array} \right] \label{eq:Aestpart}}    
The reconciled estimates are obtained using the estimated constraint matrix in \eqrefn{Eq:recondata}.  The following example demonstrates how partial knowledge of constraints can be combined with PCA to obtain improved estimates of variables as well as error-covariance matrix.

\textit{Example 6} \\ The flow system of Example 1 is considered here.  It is assumed that the following constraints are known
\bea{\Abf_g=\bMat{1 & 1 & -1 &  0 & 0 &0\\0& 0& 1& -1& 0& 0}.} However, the error-covariances, and the number of constraints are unknown. IPCA is applied to the data matrix to estimate the measurement error variances and remaining constraints.  The six singular values obtained at convergence are 258.17, 21.51, 1.01, 0.99, 0, and 0. The eigenvectors corresponding to the last two zero singular values correspond to the given constraints. Two singular values are equal to one, and, hence two additional constraints are correctly identified. The standard deviations of measurement errors estimated using IPCA and the RMSE in reconciled estimates are shown in Table~\ref{Table:Example6}.  Comparing with the results reported in Table~\ref{Tab:Example4}, it can be inferred that by utilizing knowledge of the two known process constraints, IPCA is able to obtain slightly more accurate reconciled estimates. 
 \begin{table}[htdp]
 \caption{Partially known constraint matrix: Standard deviations (SD) and root mean square errors in the reconciled estimates of flow variables using IPCA (Example 6) }
 \begin{center}
 
 \begin{tabular}{lll} \hline
 Variable & Estimated SD& RMSE using IPCA  \\ \hline \hline
 $F_1$&0.1007 & 2.3418\\\hline
 $F_2$& 0.0850 & 1.6689 \\\hline
 $F_3$&  0.1513& 2.5189\\\hline
 $F_4$&  0.2128& 2.5189\\\hline
 $F_5$&  0.1872& 2.3342\\\hline
 $F_6$&  0.0865& 1.6692\\\hline
 \end{tabular} \label{Table:Example6}
 \end{center}

 \end{table}
 The subspace angle between the row spaces of estimated and true constraint matrices is 0.0188 degree and $\alpha$ = 0.0032.  Further, the estimated regression matrix  computed by considering the flow variables $F_3$ to $F_6$ as dependent variables is as follows:
 
  \bea{
   \hat{\Rbf}_{par}&=& \bMat{1.0000 & 1.0000\\1.0000& 1.0000\\0.9999 & -0.0007\\0.0002& 1.0003}.}
   The maximum absolute difference between the elements of $\Rbf$ and $\hat{\Rbf}_{par}$ is  7.4346 $\times$ 10$^{-4}$.  These values indicate that the incorporation of the given constraints into IPCA  (PCA) method improves the estimation of the remaining constraints.
\subsection{Selection of model order}\label{Section:modelorder}
If the assumptions made regarding the measurement errors hold in practice, and the data are obtained when the process is strictly in a steady state, then the model order can be determined by examination of the singular values as described in Sections \ref{sec:ModelID} and \ref{sec:IPCA}.  If these assumptions do not hold, then it is difficult to precisely identify the model order from data when PCA or IPCA is applied.  In such a scenario, the estimated model order $m_e$ may be either greater or less than the true model order $m$.  If the estimated model order is greater than the true model order, we refer to it as \emph{overfitting} because the reconciled estimates are forced to satisfy additional constraints that are not valid.  Conversely, if the estimated number of constraints is less than the actual number, then we refer to it as \emph{underfitting}. In this section, we investigate the effect of underfitting and overfitting the data, due to incorrect estimation of the number of constraints.

We write the SVD of the data matrix in partitioned form as given by Eq. \eqref{Eq:denoising}, where $\Ubf_2$ contain the eigenvectors of $\Sbf_y$ corresponding to the smallest $m$ eigenvalues.  In the case of underfitting, the eigenvectors corresponding to the smallest $m_e < m$ eigenvalues, denoted as $\Ubf_{2s}$ will be chosen.  This implies that $\Ubf_{2s}$ is a subset of $\Ubf_2$, and the columns of $\Ubf_{2s}$ will therefore be an  asymptotically unbiased estimate of a subset of linear combinations of the rows of $\Abf$.  Therefore, in the limit as sample size goes to infinity, the estimated constraint matrix will be orthogonal to the true values of the variables.  That is,
\be{\Ubf_{2s}^{\tr}\Xbf = \zeros \label{eq:orthoprop}}
The reconciled estimates corresponding to the identified constraint matrix is given by
\bea{\hat{\Xbf} &=& \Ybf - \Ubf_{2s}\Sbf_{2s}\Vbf_{2s}^{\tr} \\
&=&  (\Ibf - \Ubf_{2s}\Ubf_{2s}^{\tr})\Ybf. \label{eq:underfitxest}}
Taking expectation of the above equation and using Eq. \eqref{eq:orthoprop} we can prove that
\be{\Exp{\hat{\Xbf}} = \Xbf. \label{eq:underfitbias}}
\eqrefn{eq:underfitbias} indicates that the reconciled estimates are unbiased even if the estimated number of constraints is less than the actual number.

If the number of constraints estimated is greater than $m$, then the estimated constraint matrix is given by
\be{\hat{\Abf} =  \left[ \begin{array}{l} \Ubf_2^{\tr} \\ \Ubf_{1s}^{\tr}  \end{array} \right] \label{eq:overfitAest}}
where $\Ubf_{1s}$ is a subset of $\Ubf_1$.  In the limit as the sample size goes to infinity, the eigenvectors $\Ubf_2$ will be orthogonal to the true values, but the eigenvectors $\Ubf_{1s}$ form a subspace of the true data space.  This implies that
\be{\left[ \begin{array}{l} \Ubf_2^{\tr} \\ \Ubf_{1s}^{\tr}\end{array} \right] \Xbf = \left[ \begin{array}{l} \zeros \\ \alpha \end{array} \right] \label{eq:overfitbias}}
Substituting the above estimate of the constraint matrix in \eqrefn{eq:reconest}, the reconciled estimates are obtained as
\bea{\hat{\Xbf} & = &\Ybf - [\Ubf_2 \ \Ubf_{1s}]\left[ \begin{array}{l} \Ubf_2^{\tr} \\ \Ubf_{1s}^{\tr}  \end{array} \right]\Ybf \\ & = & \Ybf - (\Ubf_2\Ubf_2^{\tr} + \Ubf_{1s}\Ubf_{1s}^{\tr})\Ybf}
Taking the expectation of above equation and using \eqrefn{eq:overfitbias} we get
\be{\Exp{\hat{\Xbf}} = \Xbf - \Ubf_{1s}\alpha \label{eq:overfitbiasx}}
This shows that overfitting introduces a bias in the reconciled values. Based on the above analysis, we make the recommendation that in practice it is better to be conservative in estimating the number of constraints, and avoid overfitting when using heuristics to determine the model order.  The following example demonstrates the effect of underfitting and overfitting the data. 

\textit{Example 7} \\
In this example, we will demonstrate effect of selection of model order on the estimates of the constraints, standard deviations, and the reconciled values. The  settings  of Example 4 are considered to demonstrate the concept.  IPCA is applied to the data matrix under the assumption of different choices of model order.  The singular values obtained at convergence of the method are reported for different model orders in Table~\ref{Table:SVH}. Examination of the singular values show that if the model order assumed is $5$ which is greater than the true model order of $4$, then the last five singular values are not all equal to unity. This clearly indicates that the assumed model order is incorrect.  However, if the assumed model order is equal to or less than the true model order, then the number of unity singular values obtained is equal to the assumed model order. Based on these observations, a systematic method can be devised for determining the true model order using IPCA.  We start with the least value of the model order that results in an identifiable problem, that is the assumed value of $m$ should be such that $m(m+1)/2 \geq n$ (in this example it is $3$ since we have to estimate $6$ error variances).  If the number of unity singular values obtained at convergence is equal to $m$, then we cannot conclude that the assumed model order is correct.  Instead we increment the assumed model order by unity and apply IPCA again.  If the number of unity singular values obtained is not equal to the assumed model order, then this violates the theoretical result expected from IPCA.  The true model order is one less that the assumed model order at which this violation is observed. Although, in theory the true model order can be estimated exactly using this systematic procedure with IPCA, it may not work in practice if the noise in measurements do not satisfy the assumptions made, or there is mild nonlinearity in the process constraints which we have not considered.  In practical applications when there is an ambiguity in selecting the model order it is better to choose a lower value.  This is clearly shown by the RMSE values of the reconciled estimates obtained for this example for different model-order choices reported in Table~\ref{Table:HomoscedasticRMSE}. Comparing the RMSE values, it can be seen that overfitting leads to poor reconciled estimates due to bias introduced in the estimates. In contrast underfitting only marginally increases the inaccuracy in the reconciled estimates. The  $\alpha$ values computed for various orders in Table~\ref{Table:alphaipca} also indicate that the overfitting leads to poor estimates of the constraint model. 




 \begin{table*}
 \caption{Singular values for different model order selection  }
 \begin{center}
 \begin{tabular}{lllll} \hline
No. &  \multicolumn{3}{c}{Singular values }   \\
& \multicolumn{3}{c}{}\\  \hline \hline
  & $m_e=5$ & $m_e=4$ & $m_e=3$ \\
  &  (overfitting) &(perfect model order)& (underfitting)\\ \hline \hline
 1 &  30194.45 &  258.47  &  57191.37 \\ \hline
  2 & 1.98 & 21.53  & 41.35\\ \hline
 3 & 1.00  & 1.02   &   2.54 \\ \hline
 4 & 0.16 &   0.99 &  1.01 \\ \hline
  5 & 0.11     &  0.99  &   0.99 \\ \hline
  6  & 0.07  &   0.99  &  0.99 \\ \hline
 \end{tabular}
 \end{center}
 \label{Table:SVH}
 \end{table*}

 \begin{table*}
 \caption{RMSE values for different model order selection for each flow variable }
 \begin{center}
 \begin{tabular}{lllll} \hline
 Variable &  \multicolumn{3}{c}{RMSE values }   \\
& \multicolumn{3}{c}{}\\  \hline \hline
  & $m_e=5$ & $m_e=4$ & $m_e=3$ \\
  &  (overfitting) &(perfect model order)& (underfitting)\\ \hline \hline
 $F_1$ & 35.4270 &  2.3439  &   2.6043 \\ \hline
  $F_2$ & 35.5942 & 1.6753  &1.6943 \\ \hline
 $F_3$ & 4.8291  & 2.5246   &   2.6577 \\ \hline
 $F_4$ & 4.8345 &   2.5309 &  2.6314 \\ \hline
  $F_5$ & 35.4262     &  2.3344  &    5.8500 \\ \hline
  $F_6$  & 35.5942  &   1.6689  &  1.6766 \\ \hline
 \end{tabular}
 \end{center}
 \label{Table:HomoscedasticRMSE}
 \end{table*}

\begin{table*}
\caption{Quality of the model identified for different model orders by IPCA }
\begin{center}
\begin{tabular}{ll} \hline
model order & $\alpha$\\ \hline \hline
$m_e=5$ (overfitting) & 0.9125       \\
 $m_e=4$ (true model-order)& 0.0509\\
  $m_e=3$ (underfitting) & 0.0492\\ \hline
\end{tabular} \label{Table:alphaipca}
\end{center}
\end{table*}
\section{Conclusions}\label{sec:conclusion}
PCA  has been primarily regarded as a multivariate statistical technique which is useful for reducing dimensionality of data as well as for denoising it.  The main message that we have attempted to convey in this paper is that PCA is a method that can be used to derive the steady state constraints of a linear process entirely from data along with reconciliation of the measurements.  Iterative PCA which is a recent extension of PCA derives both the error-covariance matrix of measurements and the process constraint model simultaneously from data.  Thus, the information necessary to reconcile data, and the reconciled estimates can be extracted from the data itself without any a priori process knowledge using PCA (or IPCA). We have also shown that if partial knowledge of process constraints is available, then they can be exploited to improve the estimates. Further, the effect of model order on the reconciled values  have been studied, and it is shown that it is better to be conservative in fitting the number of constraints (i.e. estimate less constraints) in case of unknown model order.

\begin{figure}
\begin{center}
\includegraphics[scale=0.5]{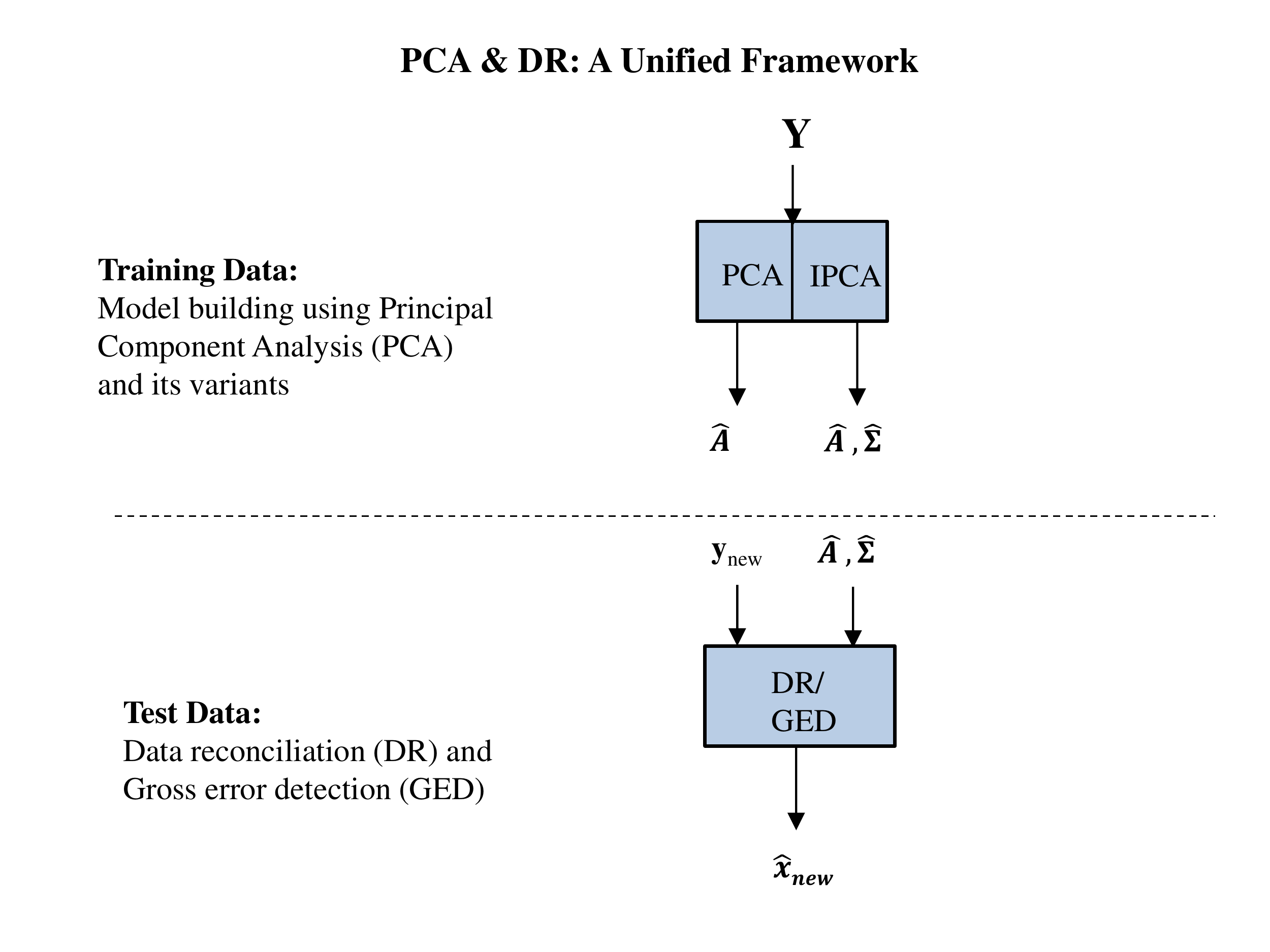}
\end{center}
\caption{A unified framework for principal component analysis (PCA) and data reconciliation (DR)} \label{PCADRFramework}
\end{figure}

 The perspective provided in this paper can be used to seamlessly integrate PCA or IPCA with data reconciliation (DR) and its companion technique of gross error detection (GED). This integration is depicted in Fig.~\ref{PCADRFramework}.  We propose that IPCA should be used on historical (training) data to derive the process model and error-covariance matrix, if  they cannot be easily obtained using \emph{a priori} process  knowledge.  The derived model and error-covariance matrix can be used in the techniques of DR and GED for reconciling future measurements and to detect gross errors in these measurements or the model as required.

\section*{Acknowledgment}
 The financial support to Dr.~Nirav Bhatt from Department of Science $\&$ Technology, India through INSPIRE Faculty Fellowship is acknowledged. 
\bibliography{biblio}

\begin{thebibliography}{18}
\expandafter\ifx\csname natexlab\endcsname\relax\def\natexlab#1{#1}\fi
\providecommand{\bibinfo}[2]{#2}
\ifx\xfnm\undefined \def\xfnm[#1]{\unskip,\space#1}\fi
\bibitem[{Almasy and Mah(1984)}]{Almasy84}
\bibinfo{author}{Almasy\xfnm[GA]}; \bibinfo{author}{Mah\xfnm[RSH]}.
\newblock \bibinfo{title}{Estimation of measurement error variances from
  process data}.
\newblock \bibinfo{journal}{Ind Eng Chem Process Des Dev}
  \bibinfo{year}{1984};\bibinfo{volume}{23}:\bibinfo{pages}{779--784}.
\bibitem[{Bagajewicz(2001)}]{Bagajewicz01}
\bibinfo{author}{Bagajewicz\xfnm[MJ]}.
\newblock \bibinfo{title}{Process Plant Instrumentation: Design and Upgrade}.
\newblock \bibinfo{publisher}{Technomic Publishing Company, Lancster},
  \bibinfo{year}{2001}.
\bibitem[{Chen et~al.(1997)Chen, Bandoni and Romagnoli}]{ChenBR97}
\bibinfo{author}{Chen\xfnm[J]}; \bibinfo{author}{Bandoni\xfnm[A]};
  \bibinfo{author}{Romagnoli\xfnm[JA]}.
\newblock \bibinfo{title}{Robust estimation of measurement error
  variance/covaraince from process sampling data}.
\newblock \bibinfo{journal}{Comput Chem Eng}
  \bibinfo{year}{1997};\bibinfo{volume}{21}:\bibinfo{pages}{593--600}.
\bibitem[{Crowe et~al.(1983)Crowe, Garcia~Campos and Hrymak}]{Crowe83}
\bibinfo{author}{Crowe\xfnm[CM]}; \bibinfo{author}{Garcia~Campos\xfnm[GY]};
  \bibinfo{author}{Hrymak\xfnm[A]}.
\newblock \bibinfo{title}{Reconciliation of process flow rates by matrix
  projection}.
\newblock \bibinfo{journal}{AIChE J}
  \bibinfo{year}{1983};\bibinfo{volume}{29}:\bibinfo{pages}{881--888}.
\bibitem[{Davis et~al.(1999)Davis, Piovoso, Kosanovich and Bakshi}]{Davies00}
\bibinfo{author}{Davis\xfnm[J]}; \bibinfo{author}{Piovoso\xfnm[MJ]};
  \bibinfo{author}{Kosanovich\xfnm[KA]}; \bibinfo{author}{Bakshi\xfnm[BR]}.
\newblock \bibinfo{title}{Process data analysis and interpretation}.
\newblock In: \bibinfo{booktitle}{Advances in chemical engineering}.
  \bibinfo{publisher}{Academic Press}; \bibinfo{year}{1999}. p.
  \bibinfo{pages}{2--97}.
\bibitem[{Gertler et~al.(1999)Gertler, Li, Huang and McAvoy}]{Gertler99}
\bibinfo{author}{Gertler\xfnm[J]}; \bibinfo{author}{Li\xfnm[W]};
  \bibinfo{author}{Huang\xfnm[Y]}; \bibinfo{author}{McAvoy\xfnm[T]}.
\newblock \bibinfo{title}{Isolation enhanced principal component analysis}.
\newblock \bibinfo{journal}{AIChE J}
  \bibinfo{year}{1999};\bibinfo{volume}{45}:\bibinfo{pages}{323--334}.
\bibitem[{Hodouin(2010)}]{Hodouin10}
\bibinfo{author}{Hodouin\xfnm[D]}.
\newblock \bibinfo{title}{Process observers and data reconciliation using mass
  and energy balances}.
\newblock In: \bibinfo{editor}{Sbarbaro\xfnm[D]};
  \bibinfo{editor}{Villar\xfnm[R]}, editors. \bibinfo{booktitle}{Advanced
  Control and Supervision of Mineral Processing Plants}.
  \bibinfo{publisher}{Springer}; \bibinfo{year}{2010}. p.
  \bibinfo{pages}{15--83}.
\bibitem[{Huang(2001)}]{Huang01}
\bibinfo{author}{Huang\xfnm[B]}.
\newblock \bibinfo{title}{Process identification based on last principal
  component analysis}.
\newblock \bibinfo{journal}{J Proc Cont}
  \bibinfo{year}{2001};\bibinfo{volume}{11}:\bibinfo{pages}{19--33}.
\bibitem[{Jolliffe(2002)}]{Jolliffe02}
\bibinfo{author}{Jolliffe\xfnm[IT]}.
\newblock \bibinfo{title}{Principal Component Analysis}.
\newblock \bibinfo{edition}{2nd} ed.
\newblock \bibinfo{publisher}{Springer-Verlay, New York}, \bibinfo{year}{2002}.
\bibitem[{Keller et~al.(1992)Keller, Zasadzinski and Darouach}]{Keller92}
\bibinfo{author}{Keller\xfnm[JY]}; \bibinfo{author}{Zasadzinski\xfnm[M]};
  \bibinfo{author}{Darouach\xfnm[M]}.
\newblock \bibinfo{title}{Analytical estimator of measurement error variances
  in data reconciliation}.
\newblock \bibinfo{journal}{Comput Chem Eng}
  \bibinfo{year}{1992};\bibinfo{volume}{16}:\bibinfo{pages}{185--188}.
\bibitem[{Kourti and MacGregor(1995)}]{Kourti95}
\bibinfo{author}{Kourti\xfnm[T]}; \bibinfo{author}{MacGregor\xfnm[JF]}.
\newblock \bibinfo{title}{Process analysis, monitoring and diagnosis, using
  multivariate projection methods}.
\newblock \bibinfo{journal}{Chemometrics Intell Lab Syst}
  \bibinfo{year}{1995};\bibinfo{volume}{28}:\bibinfo{pages}{3--21}.
\bibitem[{Mah(1990)}]{MahR90}
\bibinfo{author}{Mah\xfnm[RSH]}.
\newblock \bibinfo{title}{Chemical Process Structures and Information Flows}.
\newblock \bibinfo{publisher}{Butterworths Publications, USA},
  \bibinfo{year}{1990}.
\bibitem[{Narasimhan and Jordache(2000)}]{NarasimhanJ00}
\bibinfo{author}{Narasimhan\xfnm[S]}; \bibinfo{author}{Jordache\xfnm[C]}.
\newblock \bibinfo{title}{Data Reconciliation \& Gross Error Detection: An
  Intelligent Use of Process Data}.
\newblock \bibinfo{publisher}{Gulf Publishing Company}, \bibinfo{year}{2000}.
\bibitem[{Narasimhan and Shah(2008)}]{NarasimhanS08}
\bibinfo{author}{Narasimhan\xfnm[S]}; \bibinfo{author}{Shah\xfnm[S]}.
\newblock \bibinfo{title}{Model identification and error covariance matrix
  estimation from noisy data using pca}.
\newblock \bibinfo{journal}{Cont Engg Prac}
  \bibinfo{year}{2008};\bibinfo{volume}{16}:\bibinfo{pages}{146--155}.
\bibitem[{Rao(1964)}]{Rao64}
\bibinfo{author}{Rao\xfnm[CR]}.
\newblock \bibinfo{title}{The use and interpretation of principal component
  analysis in applied research}.
\newblock \bibinfo{journal}{Sankhya, Ser A}
  \bibinfo{year}{1964};\bibinfo{volume}{26}:\bibinfo{pages}{329--358}.
\bibitem[{Romagnoli and Sanchez(1999)}]{romagnolibook99}
\bibinfo{author}{Romagnoli\xfnm[J]}; \bibinfo{author}{Sanchez\xfnm[M]}.
\newblock \bibinfo{title}{Data Processing and Reconciliation For Chemical
  Process Operation}.
\newblock \bibinfo{publisher}{Academic Press}, \bibinfo{year}{1999}.
\bibitem[{Veverka and Madron(1997)}]{Madron97}
\bibinfo{author}{Veverka\xfnm[VV]}; \bibinfo{author}{Madron\xfnm[F]}.
\newblock \bibinfo{title}{Material and Energy Balancing in the Process
  Industries: From Microscopic Balances to Large Plants}.
\newblock \bibinfo{publisher}{Elsevier}, \bibinfo{year}{1997}.
\bibitem[{Yoon and MacGregor(2001)}]{Yoon01}
\bibinfo{author}{Yoon\xfnm[S]}; \bibinfo{author}{MacGregor\xfnm[JF]}.
\newblock \bibinfo{title}{Fault diagnosis with multivariate statistical models
  part i: using steady state fault signatures}.
\newblock \bibinfo{journal}{J of Proc Cont}
  \bibinfo{year}{2001};\bibinfo{volume}{11}:\bibinfo{pages}{387--400}.

\end{thebibliography}

\end{document}